\documentclass[11pt]{article}

\usepackage[final]{acl}

\usepackage{times}
\usepackage{latexsym}

\usepackage[T1]{fontenc}

\usepackage[utf8]{inputenc}

\usepackage{microtype}

\usepackage{graphicx}

\usepackage{amssymb}
\usepackage{amsmath}

\usepackage{booktabs}
\usepackage{multirow}
\usepackage{dblfloatfix}

\usepackage[skip=5pt]{caption}

\title{Post-edits Are Preferences Too}

\author{Nathaniel Berger$^{\ast}$, Miriam Exel$^\ddag$, Matthias Huck$^\ddag$ \and Stefan Riezler$^{\dagger,\ast}$\\ 
  $^{\ast}$Computational Linguistics \& $^\dagger$IWR, Heidelberg University, Germany \\
  $^\ddag$SAP SE, Dietmar-Hopp-Allee 16, 69190 Walldorf, Germany\\
  {\tt \{berger, riezler\}@cl.uni-heidelberg.de}\\  
  {\tt \{miriam.exel, matthias.huck\}@sap.com}}

\begin{document}
\maketitle
\begin{abstract}
Preference Optimization (PO) techniques are currently one of the state of the art techniques for fine-tuning large language models (LLMs) on pairwise preference feedback from human annotators. However, in machine translation, this sort of feedback can be difficult to solicit. Additionally, \citet{kreutzer-etal-2018-reliability} have shown that, for machine translation, pairwise preferences are less reliable than other forms of human feedback, such as 5-point ratings.

We examine post-edits to see if they can be a source of reliable human preferences by construction. In PO, a human annotator is shown sequences $s_1$ and $s_2$ and asked for a preference judgment, 
while for post-editing, editors \emph{create} $s_1$ and know that it should be better than $s_2$. We attempt to use these implicit preferences for PO and show that it helps the model move towards post-edit-like hypotheses and away from machine translation-like hypotheses. Furthermore, we show that best results are obtained by pre-training the model with supervised fine-tuning (SFT) on post-edits in order to promote post-edit-like hypotheses to the top output ranks.
\end{abstract}

\section{Introduction}

The current state of the art methods for training large language models offline on human preference data are Direct Preference Optimization (DPO) \citep{dpo_paper} or Identity Preference Optimization (IPO) \citep{ipo_paper}. Instead of training a separate reward model and then performing reinforcement learning, these methods train directly on the collected preference data by deriving a directly optimizable loss function from the preference model. 

However, in some domains, the pairwise preference annotations required for using these methods have been found to be less reliable than other annotation schemes. \citet{kreutzer-etal-2018-reliability} find that inter-rater reliability for pairwise ranking of machine translation outputs to be less than that of 5-point rating. In the field of translation, there are many different dimensions on which one translation may be better than another, e.g. fluency, faithfulness, formality, terminology, etc. \citep{lommel-2013-multidimensional}. This poses a problem for human annotators when they are presented with two plausible translations. 

\begin{figure}[t!]
    \centering
    \includegraphics[width=0.25\textwidth]{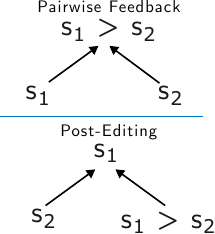}
    \caption{The generative process for preference optimization is that two sequences $s_1$ and $s_2$ are given, and a preference judgment $s_1 > s_2$ is generated (upper graph). The data generating process of post-editing yields reliable preferences by construction: Given $s_2$ and the implicit preference that $s_1 > s_2$, create $s_1$ (lower graph). We propose using the implicit preferences from post-editing for preference optimization.}
    \label{fig:generative_processes}
\end{figure}

We propose using the data generated by post-editing to yield reliable preferences by construction. The current generative process for preference data is that two sequences $s_1$ and $s_2$ are given, and a preference judgment $s_1 > s_2$ is sought, yielding the generative process $s_1 \rightarrow s_1 > s_2 \leftarrow s_2$. We propose using data generated by the following process: Given $s_2$ and the implicit preference that $s_1 > s_2$, create $s_1$, yielding the generative process $s_1 > s_2  \rightarrow s_1  \leftarrow s_2$
(see Fig. \ref{fig:generative_processes}).

Post-editing is already a common practice in the translation community to clean up raw-MT outputs before publishing. Post-editors create new sequences that they prefer with regards to the quality expected in their domain. Typically, the original raw-MT output is discarded and the post-edit is published. If this data is used for training, the post-edit is treated as a new reference for supervised fine-tuning (SFT). This ignores, however, the fact that the post-edit is not just a new reference translation but also a quality judgment of what in the raw-MT was erroneous. Using PO objectives allows us to fine-tune an LLM to translate in a way that is more in line with the post-editors' implicit preferences. However, PO does not necessarily promote the preferred sequence to become the argmax output of the model, but rather re-ranks sequences within the model's probability space.  If the two sequences are both unlikely under the model's output distribution, they will remain unlikely but their relative probability will respect the preferences. We show that best results are obtained by pre-training the model on post-edits with SFT, promoting post-edits to the top ranks, followed by fine-tuning with a PO loss.
This combined training teaches the model to prefer and promote post-edits such that reference-like translations are produced but also dispreferred machine-translation-like hypotheses are avoided.


\section{Related Work}

\citet{kreutzer-etal-2018-reliability} gather human feedback on machine translation outputs in the form of 5-point ratings and as pairwise preferences. They then use this feedback to train two reward models, one that is trained on the 5-point ratings and is trained with a regression loss to directly predict a reward value and one that is trained on pairwise preferences by fitting a Bradley-Terry model \citep{bradley1952rank} to the preferences as had been done by \citet{deep_rl_hf}. These reward models are then used to train machine translation models. \citet{kreutzer-etal-2018-reliability} find that ratings are more reliable than rankings and that reinforcement learning with a ratings-trained reward estimator yields better results than using rankings-trained reward estimates. 

\citet{berger-etal-2023-enhancing} fine-tune a pre-trained NMT model on post-editing data by presenting the model with both the post-edit and the current MT hypothesis. At each epoch, the NMT model being trained generates translations for all training data. These generated outputs are then compared to the original post-edits with a token-level diff. Both sequences are then used as training examples for the NMT system. However, tokens that appear in the hypothesis but not the post-edit are given a negative weight in the loss function. On examples where the two sequences differ, the model gets both negative feedback, where the probability of that token is to be decreased, and positive feedback, where the probability should be increased. 

\citet{xu2024contrastive} similarly present the model with a positive and negative example of machine translation outputs during training but use a modified version of the DPO \citep{dpo_paper} loss to optimize it. Their change to DPO adds an SFT term. The SFT term promotes the preferred sequence to be the argmax output of the model while the DPO part of the loss establishes the distance between the two sequences in log-probability space. The MT hypotheses that they generate come from two different LLMs; ALMA-13B-LoRA \citep{xu2024paradigmshiftmachinetranslation} and GPT-4 \citep{openai2024gpt4technicalreport}. Additionally, they use reference translations from the original dataset. The preferences that they use are predicted by open-source quality estimation models KIWI-XXL \citep{rei-etal-2023-scaling} and XCOMET \citep{XCOMET_paper}.  

\section{Preference Optimization}

\subsection{Background}

Using reinforcement learning with human feedback (RLHF) has recently re-emerged as a method for training LLMs to generate outputs that are preferred by human annotators (\citet{ZieglerETAL:19}, \citet{instructgpt}, \citet{bai2022traininghelpfulharmlessassistant}, inter alia) without requiring handwritten demonstrations of preferred behavior which would be required for supervised fine-tuning (SFT). The general recipe is as follows: pre-train an LLM on in-domain data; generate multiple completions $y$ for a single input $x$ (or prompt); have human annotators rank or rate the completions; train a reward model to predict rankings or ratings given inputs and completions; use the trained reward model to predict rewards for reinforcement learning, frequently with proximal policy optimization \citep{schulman2017proximal}. Training a separate model to predict rewards for reinforcement learning is known as an actor-critic method.

The reward model in the previous works is structured as a Bradley-Terry model \citep{bradley1952rank}, where the probability of preferring $y_1$ over $y_2$ is given by 
$$
p(y_1 \succ y_2 | x) = \sigma(r_\theta(x,y_1) - r_\theta(x, y_2))
$$
where $\sigma$ is the logistic function and $r_\theta$ is the reward model that is trained on the pairwise rankings to give the preferred sequence a higher value. The reward model can then be used to estimate rewards for outputs sampled during online training.  

This process requires training an additional model and hiring human annotators to perform ranking. DPO \citep{dpo_paper} is a technique that obviates the need for a secondary model by instead giving the model both the preferred and dispreferred sequences and optimizing a distance between the two sequence in log-probability space. 

If handwritten demonstrations of preferred sequences are available, then SFT would typically be performed. The goal of SFT is to maximize the probability of the demonstrations under the model. For text generation, this is done by minimizing the negative log-probability of each token given all previous tokens in the sequence. 
$$\mathcal{L}_{SFT}(y) = - \sum_{i=0}^{|y|} \log(\pi(y_i|y_{0:i-1}))$$
Minimizing this loss promotes the sequence $y$ to be the argmax output of the model, while reinforcement learning increases or decreases the probability of a sequence with regard to the magnitude of its reward.

\subsection{PO Objectives}

The DPO loss \citep{dpo_paper} is based on the Bradley-Terry model of human preferences but, unlike actor-critic reinforcement learning techniques, it does not train a separate reward model.  Instead they rewrite the reward function $r$ in terms of the optimal policy and the baseline model. They notice that the theoretically optimal policy $\pi_r$, with a KL-divergence constraint, is equal to the baseline model with its output distribution re-weighted according to the reward function
$$
\pi_r(y|x) = \frac{1}{Z(x)}\pi_{ref}(y|x)\exp\left(\frac{1}{\beta}r(x,y)\right)
$$
where $Z$ is the partition function, which normalizes the function to be a proper probability distribution. This formula can also be solved for the reward function $r$, such that rewards are expressed as the difference between two models' probability ratios. If this $r$ is then inserted back into the Bradley-Terry model, it becomes 
\begin{multline*}
p(y_w \succ y_l|x) = \\
\sigma\left(\beta\log\left(\frac{\pi^*(y_w|x)}{\pi_{ref}(y_w|x)}\right) - \beta\log\left(\frac{\pi^*(y_l|x)}{\pi_{ref}(y_l|x)}\right)\right)
\end{multline*}
where $y_w$ and $y_l$ denote the preferred and dispreferred completion, respectively, and $\pi^*$ is now the model we are training to be optimal under the reward function. This probability, $p(y_w \succ y_l|x)$ can be optimized by minimizing the negative log-probability.
With regards to output probability, the loss monotonically decreases as $y_w$ becomes more probable than $y_l$. Increasing the difference between the two always decreases the loss.

\citet{ipo_paper} re-derive a similar loss with some theoretical advantages. Instead of training the model to be optimal under the Bradley-Terry derived reward function, they train the model to separate the two outputs by a fixed difference in log-probability space.
\begin{multline*}
\mathcal{L}_{IPO}(y_w, y_l, x) = \\ 
+ \left(\!\left(\!\log\!\left(\!\frac{\pi^*(y_w|x)}{\pi_{ref}(y_w|x)}\!\right) \!- \!\log\!\left(\!\frac{\pi^*(y_l|x)}{\pi_{ref}(y_l|x)}\!\right)\!\right) \!-\!\frac{1}{2\beta}\!\right)^2
\end{multline*}
Because this loss function is minimized when the log-probability ratio difference is exactly $(2\beta)^{-1}$, and will increase when the outputs move further apart in log-probability space, the authors claim an advantage for deterministic preferences, where the same preferences are seen multiple times during training. Because the preferences that we use are deterministic, we opt for the IPO paradigm of PO.

A follow-up work to DPO, focused specifically on machine translation, additively combines the DPO loss and the SFT loss \citep{xu2024contrastive}, which the authors call Contrastive Preference Optimization (CPO). Additionally, they perform an ad-hoc modification of the DPO loss by dropping the normalizer $\pi_{ref}$ in the DPO loss so as to not perform a second forward pass on the reference model.
\begin{multline*}
\mathcal{L}_{CPO}(y_w, y_l, x) = - \log(\pi^*(y_w|x)) \\
- \log(\sigma(\beta\log(\pi^*(y_w|x)) - \beta\log(\pi^*(y_l|x))))
\end{multline*}

We use a reformulation of the CPO loss with the IPO training objective for our experiments because our preferences are deterministic. Additionally, we keep the normalizers in the IPO loss because these can be pre-computed in advance, instead of in a second forward pass, and incur only a negligible memory and speed penalty.

Our modified variant of the CPO loss, which we call dDPO for the deterministic preferences involved in post-editing, is 
\begin{multline*}
\mathcal{L}_{dCPO}(y_w, y_l, x) = - \log(\pi^*(y_w|x)) \\
+ \left(\!\left(\!\log\!\left(\!\frac{\pi^*(y_w|x)}{\pi_{ref}(y_w|x)}\!\right) \!- \!\log\!\left(\!\frac{\pi^*(y_l|x)}{\pi_{ref}(y_l|x)}\!\right)\!\right) \!-\!\frac{1}{2\beta}\!\right)^2
\end{multline*}
which is the SFT objective added to the IPO objective. When we refer to dCPO later in this paper, we are referring to this modified version of the CPO objective.

\section{Experiments}

\begin{table}[h]
    \centering
    \begin{tabular}{l|l|l|l|l}
       Data & Split & BLEU & TER & CHRF  \\
        \midrule
        \multirow{3}{*}{En$\rightarrow$DE} & Train & 49.4 & 37.6 & 71.6 \\
        & Dev & 50.9 & 36.5 & 72.5 \\
        & Test & 50.8 & 36.4 & 72.8 \\
        \midrule
        \multirow{3}{*}{En$\rightarrow$Ru} & Train & 80.9 & 13.6 & 89.9 \\
        & Dev & 80.2 & 14.9 & 89.0 \\
        & Test & 76.3 & 17.4 & 87.2 \\     
    \end{tabular}
    \caption{Token level metrics comparing the WMT APE datasets' machine translations to the post-edits.}
    \label{tab:wmt_token_metrics}
\end{table}


 \begin{table*}[!t]
\centering
\begin{tabular}{ll|ll|ll}
\toprule
        \multicolumn{6}{c}{Without References} \\
        \midrule
        &  & \multicolumn{2}{c|}{En$\rightarrow$DE} & \multicolumn{2}{c}{En$\rightarrow$Ru} \\
        & Model      & XCOMET-XL & XCOMET-XXL & XCOMET-XL & XCOMET-XXL \\
\midrule
    a & APE Data MT         & $92.78$           & $94.47$     & $93.07^{ce}$      & $91.35^{ce}$      \\
    b & APE Data PE         & $95.55^{ac}$      & $97.01^{ac}$     & $95.29^{ace}$     & $93.78^{ace}$      \\
    \midrule
    c & Tower Base          & $94.33^{a}$       & $94.75$     & $85.50$           & $65.07$      \\
    d & SFT                 & $95.63^{ac}$      & $97.01^{ac}$     & $95.29^{ace}$     & $93.55^{ace}$      \\
    e & IPO                 & $95.87^{ac}$      & $97.18^{ac}$     & $89.65^{c}$       & $72.90^{c}$      \\
    f & dCPO                 & $95.67^{ac}$      & $97.51^{abcde}$     & $95.55^{ace}$     & $93.73^{ace}$      \\
    g & SFT$\rightarrow$IPO & $95.87^{abcd}$    & $97.48^{abcde}$     & $95.62^{acde}$    & $94.40^{abcdef}$      \\
    h & SFT$\rightarrow$dCPO & $95.91^{abcdef}$  & $97.57^{abcde}$     & $95.85^{abcdefg}$ & $94.76^{abcdefg}$      \\
\midrule
\multicolumn{6}{c}{With References} \\
\midrule
        &  & \multicolumn{2}{c|}{En$\rightarrow$DE} & \multicolumn{2}{c}{En$\rightarrow$Ru} \\
        & Model      & XCOMET-XL & XCOMET-XXL & XCOMET-XL & XCOMET-XXL \\
\midrule
    a & APE Data MT         & $92.80$           & $94.20$           & $94.99^{bd}$      & $92.68^{bd}$      \\
    \midrule
    b & Tower Base          & $93.90^{a}$       & $94.44$           & $83.65$           & $65.48$      \\
    c & SFT                 & $95.57^{ab}$      & $96.77^{ab}$      & $95.36^{bd}$      & $93.30^{abd}$      \\
    d & IPO                 & $95.56^{ab}$      & $96.85^{ab}$      & $88.15^{b}$       & $72.64^{b}$      \\
    e & dCPO                 & $95.67^{ab}$      & $97.20^{abcd}$    & $95.36^{bd}$      & $93.06^{bd}$      \\
    f & SFT$\rightarrow$IPO & $95.94^{abcde}$   & $97.31^{abcd}$    & $95.77^{abcde}$   & $93.91^{abcde}$      \\
    g & SFT$\rightarrow$dCPO & $96.00^{abcde}$   & $97.36^{abcd}$    & $96.11^{abcdef}$  & $94.14^{abcde}$      \\
\bottomrule
\end{tabular}
\caption{XCOMET-XL and -XXL on the WMT 2020 En->DE and 2019 En->Ru test sets. Higher values are better. Superscripts indicate which system the given line is significantly better than with $\alpha < 0.05$ according to pair-wise bootstrap resampling. We see that initializing with an SFT model and then performing PO yields the best results.}
\label{tab:wmt_ende_enru_test_results}
\end{table*}

\begin{table*}[h!]
\centering
\begin{tabular}{l|ll|ll}
\toprule
& \multicolumn{2}{c|}{En$\rightarrow$De} & \multicolumn{2}{c}{En$\rightarrow$Ru} \\ 
Model & w Ref & w/o Ref & w Ref & w/o Ref \\ \midrule
Tower Base & 1.1396 & 1.4224 & 4.1858 & 8.1576 \\ 
SFT & 0.9174 & 1.1757 & 1.2706 & 1.4246 \\ 
IPO & 0.8240 & 0.9484 & 3.0420 & 5.3263 \\ 
dCPO & 0.8286 & 1.0092 & 1.2873 & 1.4575 \\ 
SFT$\rightarrow$IPO & 0.7985 & 0.9476 & 1.1554 & 1.3335 \\ 
SFT$\rightarrow$dCPO & 0.7978 & 0.9558 & 1.1110 & 1.2607 \\ 
\bottomrule
\end{tabular}
\caption{MetricX 23 XL results with and without references on the WMT 2020 En->DE and 2019 En->Ru test sets. Lower values are better. Results appear in line with XCOMET-XL and -XXL and reinforce our previous results.}
\label{tab:wmt_ende_enru_test_results_metricx}
\end{table*}

We fine-tune an LLM for the task of machine translation under five different conditions: SFT, IPO, dCPO, and pre-training with SFT followed by either IPO or dCPO, denoted as SFT$\rightarrow$IPO and SFT$\rightarrow$dCPO, respectively, and evaluate with the neural metrics XCOMET \citep{XCOMET_paper} and MetricX \citep{metricx_paper}. \citet{dpo_paper} pre-train their large language models on in-domain data such that they are already able to perform the requested task to begin with. For the task {single-turn dialogue}, they use the Anthropic Helpful and Harmless dialogue dataset but because no pre-training data is available, they perform SFT on the helpful answers as a pre-training step. This is similar to our conditions SFT$\rightarrow$IPO and SFT$\rightarrow$dCPO.

The LLM that we choose to fine-tune is Tower-Base by \citet{tower_llm_2024}. We make this choice because it is a multi-lingual LLM pre-trained on all languages we intend to work with and because the size of the model is still small enough to perform a full fine-tune with our resources\footnote{We train on a server with 4x Nvidia A40 GPUs with 48 GB of memory each. The system contains 256GB of RAM and 64 CPU cores.}. We opt for Tower-Base instead of Tower-Instruct because Tower-Instruct has been instruction fine-tuned for various down-stream tasks and not just for machine translation. Using Tower-Base instead allows us to perform a SFT step on our own. We fine-tune in all scenarios with a minimal prompt "Translate English to German.\textbackslash nEnglish: \{Source\}\textbackslash nGerman:" for our German examples. Our Russian examples use a prompt with the language name changed. This prompt is used for both SFT and PO training objectives.

Our post-edits come from WMT Automatic Post-Editing (APE) shared tasks of previous years. These datasets contain triples of source, machine-translation (MT), and post-edit (PE). We focus on the language pairs En$\rightarrow$De from 2020 and En$\rightarrow$Ru from 2019. The En$\rightarrow$De source data comes from Wikipedia and is translated by a black-box NMT system \citep{wmt-ape-2020-findings}. The En$\rightarrow$Ru data comes from the information technology domain from Microsoft Office localization work and was translated by Microsoft's production NMT system \citep{wmt-ape-2019-findings}. The En$\rightarrow$Ru data contains base64 encoded data and sequences long enough to cause out of memory errors. We therefore filter out sequences with fewer than 4 tokens, more than 128 tokens, or more than 500 characters from the En$\rightarrow$Ru training data, leaving 9290 (source, mt, pe) triples for training. The En$\rightarrow$De training data was already clean and all 7000 (source, mt, pe) triples were kept for training.

Table \ref{tab:wmt_token_metrics} shows the performance of the datasets' machine translations when compared to their post-edits in terms of token based metrics, BLEU, TER, and CHRF (\citet{Papineni:02}, \citet{SnoverETAL:06}, and \citet{Popovic:15}). We see that more edits were made to the German machine translations compared to the Russian machine translations. The Russian data has far more unedited sequences---of the 9290 examples we have in our Russian training data after filtering, 5263 are unedited or 56.7\%. To compare, of the 7000 German training examples, 448 are unedited or just 6.4\%. We keep the unedited data for training as the SFT and dCPO objectives can still take advantage of unedited data, but filter it out for our analysis later as it is impossible for a model to prefer an unedited "post-edit" over the machine translation.

We train with fully-sharded data parallelism (FSDP) in PyTorch using Accelerate \citep{accelerate} across four GPUs with an effective batch size of 256 sequences. When possible, we shared hyperparameters across all runs and datasets. For example, both IPO and dCPO have $\beta$ set to $0.1$. Full hyper-parameters can be found in the Appendix \ref{sec:hyperparameters}. We used reference-free XCOMET-XL as an early stopping criterion, which was run at the end of each epoch. During generation, we used greedy decoding. 

Because PO techniques requires seeing both the preferred and the dis-preferred sequence during the same optimization step, we concatenate them together along the batch dimension so that both sequences are processed in the same forward pass. This doubling of sequences in each batch requires that the number of training examples per batch be halved and the number of gradient accumulation steps be doubled in order to have the same effective batch size. This incurs no additional memory penalty but doubles the time to see the same number of training examples.

Our initial experiments showed that string-based metrics actually decrease when using PO techniques but we did not observe a discernible quality difference. This is in line with the observations reported by \cite{xu2024contrastive}. Therefore, we evaluate with neural metrics so that evaluation could not be biased towards models that produce superficially similar translations. We evaluate with XCOMET-XL and -XXL \citep{XCOMET_paper} and MetricX 23 \citep{metricx_paper}, both with and without references.

\section{Results}

\begin{figure*}[!t]
    \includegraphics[width=\textwidth]{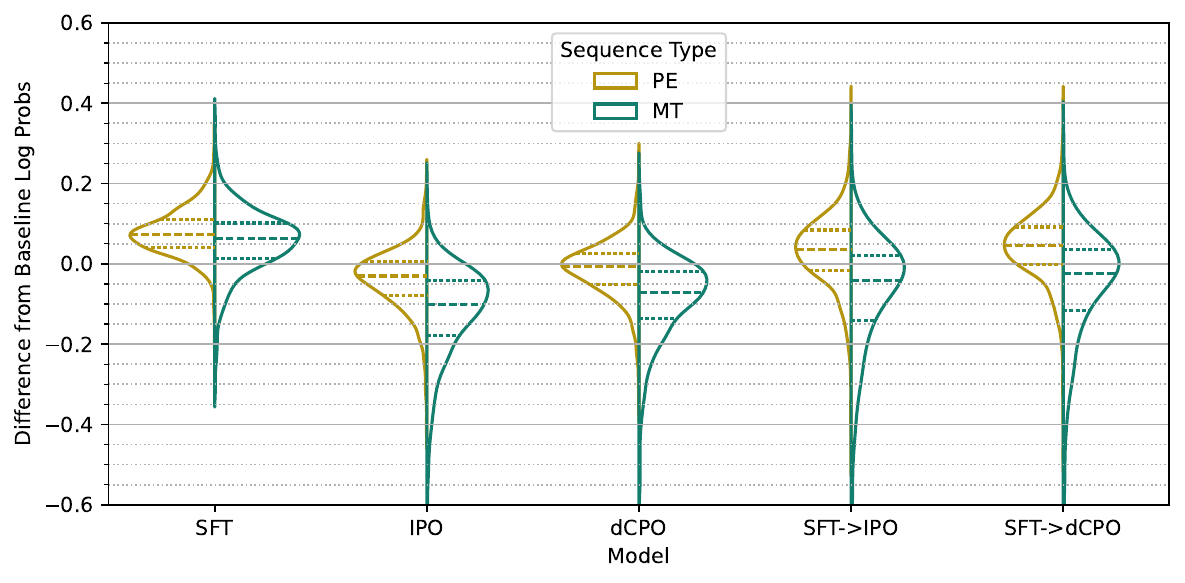}
    \caption{The difference of the models' averaged sequence log-probabilities from the baseline model's on the WMT 2020 En$\rightarrow$De test data. Zero for PE is an average log-probability of $-0.516$ while for MT it is $-0.565$. This violin plot then shows displacement from these baseline values. Dashed horizontal lines indicate quartiles.}
    \label{fig:ende_test-logp-diffs}
\end{figure*}

Our XCOMET metric results are shown in Table \ref{tab:wmt_ende_enru_test_results}. The evaluation shows that Tower Base is already competent at performing zero-shot translation for English to German, achieving reference-free XCOMET-XL and -XXL scores that are above the MT hypotheses contained in our dataset; $94.33$ and $94.75$, respectively. This is in spite of the fact that the model has not yet been instruction fine-tuned to perform zero-shot translation. 

The lack of instruction fine-tuning is made obvious in the English to Russian results, where the model is unable to translate well prior to fine-tuning. Specifically, the Tower Base model frequently translated its instructions to Russian and ignored the source text, yielding lower scores. XCOMET-XL and -XXL seem to react differently to these non-translations with XCOMET-XXL punishing them more severely than XCOMET-XL.

Supervised fine-tuning is able to reach the level of the post-edits contained in the APE datasets when evaluating with reference-free evaluation. SFT surpasses the post-edits only with XCOMET-XL on the En$\rightarrow$De data but this improvement is not significant. Here, we are evaluating the post-edits included in the dataset as if they were hypotheses for the source sentences. 

IPO and dCPO are able to improve XCOMET-XL and -XXL scores for En$\rightarrow$De above what the post-edits achieve, but only for -XXL is this improvement significant, as evaluated by pairwise bootstrap resampling implemented in the COMET package. For En$\rightarrow$Ru, only dCPO is able to surpass post-edits and even then only for XCOMET-XL. 

However, once we initialize the PO methods with the SFT model, we find our best results. SFT$\rightarrow$dCPO is significantly better than both the MT and PE data from the dataset, the Tower Base model, and the SFT model for both En$\rightarrow$De and Ru; while for just En$\rightarrow$Ru, it is better than all other systems.

Results with references do not differ drastically and can also be found in Table \ref{tab:wmt_ende_enru_test_results}. We also evaluate with MetricX 23 XL \citep{metricx_paper} and show our results in Table \ref{tab:wmt_ende_enru_test_results_metricx}. The relations follow those of XCOMET and reinforce our conclusions.

\section{Analysis}

In addition to evaluating the fine-tuned models with neural metrics, we analyze the behavior of the models after training to see how the log-probabilities of the two sequences change compared to the baseline model. Additionally, we use the log-probabilities as a measure for model preferences. If one sequence is more probable, it is preferred by the model.

In our analysis, we remove machine translation and post-edit pairs where the post-edit remains un-edited. This is because we are looking for differences in model behavior between machine translations and post-edits, which can not be done when they are the same sequence.

\subsection{Log Probability Changes}

\begin{figure*}[!t]
    \includegraphics[width=\textwidth]{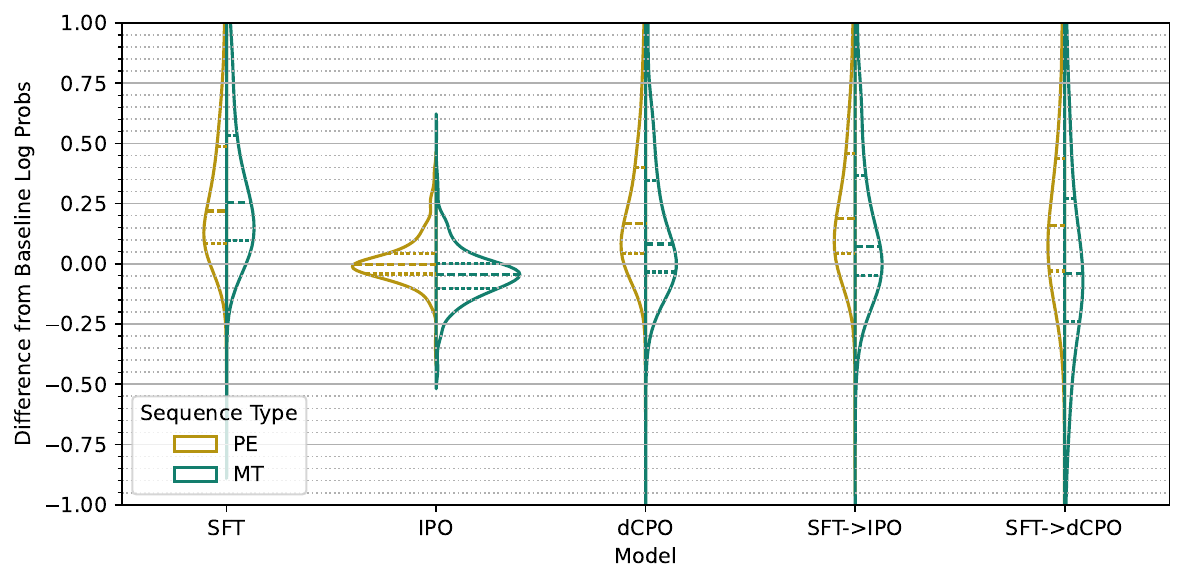}
    \caption{The difference of the models' averaged sequence log-probabilities from the baseline model's on the WMT 2019 En$\rightarrow$Ru test data. Zero for PE is an average log-probability of $-1.099$ while for MT it is $-1.260$. This violin plot then shows displacement from these baseline values. Dashed horizontal lines indicate quartiles.}
    \label{fig:enru_test-logp-diffs}
\end{figure*}

Figures \ref{fig:ende_test-logp-diffs} and \ref{fig:enru_test-logp-diffs} are split violin plots showing the difference between log-probabilities before and after training, for German and Russian respectively. The left side of each violin shows the post-edit sequences' change from the baseline model's while the right side shows the machine translations' difference. This way we can examine how each training method affects the two sequence types individually. Additionally, we also measure the difference between the post-edits and machine translations after training in Tables \ref{tab:ende_diff_between_pe_mt} and \ref{tab:enru_diff_between_pe_mt} for German and Russian, respectively.

\begin{table}
    \centering
    \begin{tabular}{l|c|c|c}
    \toprule
        & \multicolumn{3}{|c}{$PE - MT$} \\
    Model & Train & Dev & Test \\
    \midrule 
    Base & 0.038 & 0.048 & 0.049 \\
    SFT & 0.060 & 0.070 & 0.073 \\
    IPO & 0.120 & 0.124 & 0.134 \\
    dCPO & 0.110 & 0.115 & 0.124 \\
    SFT$\rightarrow$IPO & 0.144 & 0.144 & 0.157\\
    SFT$\rightarrow$dCPO & 0.138 & 0.138 & 0.150 \\
    \bottomrule
    \end{tabular}
    \caption{This table shows the average values of the post-edit log-probabilities minus the machine translation log-probabilities for the English$\rightarrow$German data. We see that the gap between PE and MT increases more with PO than it does with SFT.}
    \label{tab:ende_diff_between_pe_mt}
\end{table}

As we see in Figure \ref{fig:ende_test-logp-diffs}, if we perform SFT on post-edits, as would typically be done when treating post-edits as new references, \textit{both} the post-edits and the MT outputs become more likely under our fine-tuned model. Because the post-edits and MT outputs are highly correlated, they likely reside very close to each other in the model's hidden representation. This means, that with a smooth mapping from hidden representations to outputs, increasing the probability of the PE will also increase the probability of the MT sequence. 

For the En$\rightarrow$De IPO and dCPO runs, we see the post-edits stay close to the baseline while the MT is pushed further down in log-probability space. Additionally, the distance between the two sequences increases under PO compared to SFT. As shown in Table \ref{tab:ende_diff_between_pe_mt}, the average distance that PEs are above MT outputs doubles after PO compared to SFT. 

After the IPO training, both sequences become less likely as seen in the split violin plot for IPO in Figure \ref{fig:ende_test-logp-diffs}. This method does not have the upwards pressure on the preferred sequences that SFT or dCPO does, so we hypothesize that the downward pressure on the MT output also drags the PE sequence down as well; similar to how SFT increases the probability of MT without training on it. Alternatively, it could be that in order to establish a greater distance between the sequences, probability mass has to be re-allocated to other possible sequences.

With the En$\rightarrow$Ru data, we see that the MT sequences benefit \textit{more} from training than the PE sequences do, even though they remain unseen, as shown by the violin plot for SFT in Figure \ref{fig:enru_test-logp-diffs}. This corresponds to the smaller difference between PE and MT that we see for SFT when compared to Tower Base in Table \ref{tab:enru_diff_between_pe_mt}. The need for more fine-tuning of the Tower Base model is also visible in the SFT, SFT-initialized, and dCPO models' larger displacement from the Tower Base log-probabilities. IPO remains close to the $0$ point for both sequences because the only pressure for each sequence is for them to move further apart; which is more difficult with the large overlap between the machine-translations and post-edits.

En$\rightarrow$Ru appears similar for IPO, where both sequences are moved down in log-probability space, however the violin plot for dCPO and the SFT initialized models have a displacement from the baseline similar to SFT. This is because the baseline model was unable to perform zero-shot translation for En$\rightarrow$Ru and, since the dCPO loss includes an SFT term, it learned how to translate which moved all sequences upwards. Unlike SFT, post-edits benefit more than machine translations after dCPO training. 

Finally for the En$\rightarrow$De SFT initialized models, we see in Figure \ref{fig:ende_test-logp-diffs} that post-edits increase in probability over the baseline while machine translation outputs are held close to or below the baseline. The difference between PE and MT is increased here compared to the PO only conditions. 

We find that this behavior generalizes also to the development and test sets as shown in Table \ref{tab:ende_diff_between_pe_mt}. For En$\rightarrow$Ru, the SFT$\rightarrow$IPO model and the dCPO model both have post-edits and machine translation increase in likelihood compared to the baseline. This is again due to the baseline model being unable to perform zero-shot translation and both sequences become more likely after it is able to do so. SFT$\rightarrow$dCPO appears similar but far more stretched out and with MT moved below the baseline. This model trained for much longer before reaching its early stopping criterion (SFT$\rightarrow$dCPO stopped after 10 epochs, compared to SFT$\rightarrow$IPO stopping after 2).

The largest improvements in our XCOMET-XL and -XXL scores coincide with training methods that both move the post-edit up in log-probability space while also ensuring that the machine translations are less likely by enough of a margin. SFT on its own also increases the probability of machine translations and does not work to establish a margin between the two sequences. Additionally, this shows us that PO successfully moves the model towards generating post-edit-like translations rather than those like the machine translations.

\begin{table}
    \centering
    \begin{tabular}{l|c|c|c}
    \toprule
        & \multicolumn{3}{|c}{$PE - MT$} \\
    Model & Train & Dev & Test \\
    \midrule 
    Base & 0.039 & 0.078 & 0.161 \\
    SFT & 0.025 & 0.062 & 0.133 \\
    IPO & 0.085 & 0.125 & 0.217 \\
    dCPO & 0.101 & 0.140 & 0.240 \\
    SFT$\rightarrow$IPO & 0.110 & 0.151 & 0.263\\
    SFT$\rightarrow$dCPO & 0.192 & 0.242 & 0.419 \\
    \bottomrule
    \end{tabular}
    \caption{This table shows the average values of the post-edit log-probabilities minus the machine translation log-probabilities for the En$\rightarrow$Ru data. We see that the gap between PE and MT increases more with PO than it does with SFT.}
    \label{tab:enru_diff_between_pe_mt}
\end{table}

\subsection{Preference Changes}

\begin{figure*}[!t]
    \includegraphics[width=\textwidth]{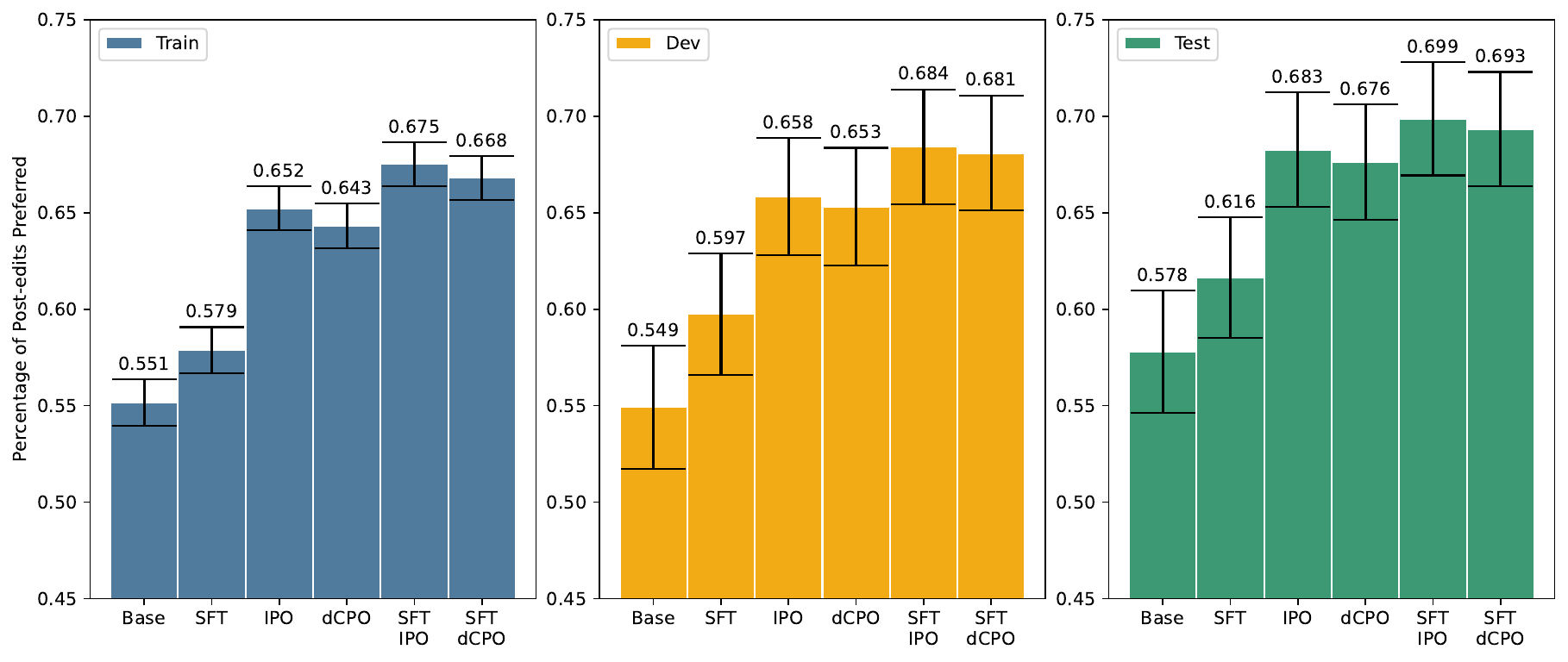}
    \caption{Here we show the percentage of training examples where the post-edit sequence is preferred in terms of average log-probability over the machine translation for the WMT En$\rightarrow$De dataset. The black lines indicate the 95\% confidence intervals for binomial distributed data---non-overlapping confidence intervals indicate a significant difference.}
    \label{fig:ende_pe_pref}
\end{figure*}

\begin{table*}[!t]
\centering
\begin{tabular}{lccc}
\toprule
\textbf{Method} & \textbf{Train} & \textbf{Dev} & \textbf{Test} \\
\midrule
Base & 55.14\% (53.94\%, 56.35\%) & 54.93\% (51.73\%, 58.12\%) & 57.80\% (54.64\%, 60.96\%) \\
SFT & 57.88\% (56.68\%, 59.07\%) & 59.74\% (56.60\%, 62.89\%) & 61.65\% (58.53\%, 64.76\%) \\
IPO & 65.23\% (64.08\%, 66.39\%) & 65.85\% (62.80\%, 68.89\%) & 68.27\% (65.29\%, 71.25\%) \\
dCPO & 64.33\% (63.17\%, 65.49\%) & 65.31\% (62.26\%, 68.36\%) & 67.63\% (64.63\%, 70.63\%) \\
SFT$\rightarrow$IPO & 67.52\% (66.39\%, 68.66\%) & 68.42\% (65.43\%, 71.40\%) & 69.87\% (66.93\%, 72.81\%) \\
SFT$\rightarrow$dCPO & 66.80\% (65.66\%, 67.94\%) & 68.09\% (65.10\%, 71.08\%) & 69.34\% (66.38\%, 72.29\%) \\
\bottomrule
\end{tabular}

\caption{Percentage of instances where post-edits are preferred over machine translations and their corresponding 95\% confidence intervals for Train, Dev, and Test Splits for the WMT En$\rightarrow$De 2020 APE Dataset. Non-overlapping confidence intervals correspond to statistically significant differences with $\alpha < 0.05$.}
\label{tab:ende_pref_table}
\end{table*}

\begin{figure*}[!t]
    \includegraphics[width=\textwidth]{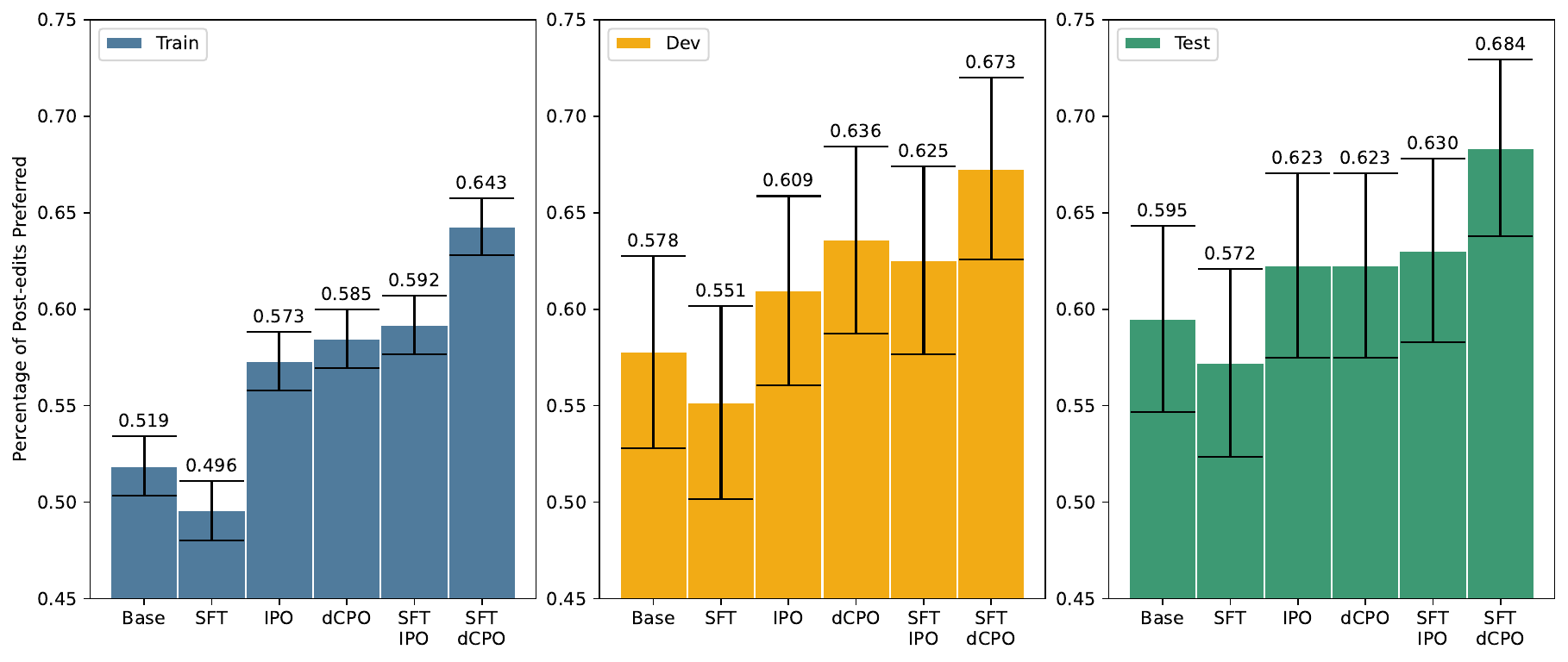}
    \caption{Here we show the percentage of training examples where the post-edit sequence is preferred in terms of average log-probability over the machine translation for the WMT En$\rightarrow$Ru dataset.. The black lines indicate the 95\% confidence intervals for binomial distributed data---non-overlapping confidence intervals indicate a significant difference.}
    \label{fig:enru_pe_pref}
\end{figure*}

\begin{table*}[!t]
\centering
\begin{tabular}{lccc}
\toprule
\textbf{Method} & \textbf{Train} & \textbf{Dev} & \textbf{Test} \\
\midrule
Base & 51.87\% (50.33\%, 53.42\%) & 57.78\% (52.81\%, 62.76\%) & 59.49\% (54.65\%, 64.33\%) \\
SFT & 49.57\% (48.02\%, 51.11\%) & 55.15\% (50.14\%, 60.15\%) & 57.22\% (52.34\%, 62.09\%) \\
IPO & 57.31\% (55.79\%, 58.84\%) & 60.95\% (56.04\%, 65.86\%) & 62.28\% (57.50\%, 67.06\%) \\
dCPO & 58.46\% (56.93\%, 59.98\%) & 63.59\% (58.74\%, 68.43\%) & 62.28\% (57.50\%, 67.06\%) \\
SFT$\rightarrow$IPO & 59.20\% (57.68\%, 60.72\%) & 62.53\% (57.66\%, 67.41\%) & 63.04\% (58.28\%, 67.80\%) \\
SFT$\rightarrow$dCPO & 64.27\% (62.79\%, 65.75\%) & 67.28\% (62.56\%, 72.01\%) & 68.35\% (63.77\%, 72.94\%) \\
\bottomrule
\end{tabular}

\caption{Percentage of instances where post-edits are preferred over machine translations and their corresponding 95\% confidence intervals for Train, Dev, and Test Splits for the WMT En$\rightarrow$Ru 2019 APE Dataset. Non-overlapping confidence intervals correspond to statistically significant differences with $\alpha < 0.05$.}
\label{tab:enru_pref_table}
\end{table*}

Changes in log-probabilities from the baseline model do not necessarily indicate whether the models' preferences have changed. It could be that, in $(mt, pe)$ pairs where it is already the case that $pe > mt$, the distances between $pe$ and $mt$ increased, but examples where $mt > pe$ did not have their ordering changed. To that end, we also examine the baseline model's preference in terms of sequence probability---if a sequence's average log probability is strictly greater than that of another sequence, it is preferred. We plot preferences across all data splits for En$\rightarrow$De in Figure  \ref{fig:ende_pe_pref} and for En$\rightarrow$Ru in Figure  \ref{fig:enru_pe_pref}. The exact values with corresponding confidence intervals are in Tables \ref{tab:ende_pref_table} and \ref{tab:enru_pref_table}, respectively.

For both language pairs, we find that the Tower Base model does not have strong preferences. On the En$\rightarrow$De data set, it prefers post-edits to machine translation 57.80\% of the time on the test set while for En$\rightarrow$Ru this preference occurs 59.49\% of the time. For En$\rightarrow$De, SFT significantly improves this preference on the training data but not on the development or test data. SFT actually seems to change the preferences in favor of machine translations on the En$\rightarrow$Ru data; which also coincides with a decrease in the average distance between sequences and machine translations increasing in probability more.

When we train with IPO and dCPO on En$\rightarrow$De, we find that both improve the preference for post-edits up to $68.27\%$ on test data. The improvements above SFT are significant for both models on the train set while for dev, the confidence intervals overlap, and for test only IPO is significantly better. On En$\rightarrow$Ru, we see a similar improvement in preferences but only on the training set are they significant. 

Initializing with SFT and then training with PO on En$\rightarrow$De yields the best improvements with $69.87\%$ on test. Both SFT$\rightarrow$IPO and SFT$\rightarrow$dCPO are significantly better than SFT across all data splits. Again, En$\rightarrow$Ru shows similar behavior with only the change on the training set being significant.

Across all data splits on En$\rightarrow$De, IPO methods seem to establish a slightly stronger preference for post-edits which seems to be accounted for by increase in difference between the two sequence types as shown in Table \ref{tab:ende_diff_between_pe_mt}. For En$\rightarrow$Ru, dCPO is better at establishing this preference which also coincides with the increase in differences from Table~\ref{tab:enru_diff_between_pe_mt}.

\section{Conclusion}

Post-editing is part of common translation workflows before publishing to clean up raw-MT outputs. If the post-edits are used for training purposes, they are treated simply as new references and the MT output is treated as a by-product. Post-edits are created with an implicit preference in mind, that the PE should be better than the MT. We find that keeping both the PE and MT allows us to perform preference optimization techniques and improve translation quality with data that would otherwise be discarded.

We find that performing supervised fine-tuning using post-edits as references also increases the likelihood of the machine translations which remained unseen by the system. However, because the original machine translations were erroneous (in order to need correction), it is disadvantageous to increase their likelihood as well. Using PO techniques allows the model to establish a larger margin between the post-edit sequence and the machine translation sequence in log-probability space.

Increasing this margin coincides with significant improvements in neural translation metrics. We additionally find that we can measure the models' preferences in terms of sequence probability---if one sequence is more likely it is preferred. Models trained with SFT do not have a significant change in preferences compared to the baseline models but using PO teaches the model to prefer the post-edits above the machine translations.

In future work, we would like to examine the effect of the distance between post-edit and machine translation sequence probabilities. Currently, IPO sets a single distance for all sequence pairs but this may be sub-optimal when the sequences are correlated to different degrees. For example, if a post-edit and machine translation share a large prefix, the rest of the tokens in the sequences must account for the distance, while for non-overlapping sequences all tokens contribute to the distance between the log-probabilities.

\section*{Acknowledgements}

The last author acknowledges support by the state of Baden-Württemberg through bwHPC
and the German Research Foundation (DFG) through grant INST 35/1597-1 FUGG.

\bibliography{custom}

\appendix

\section{Hyperparameters}
\label{sec:hyperparameters}

The hyperparameters for all runs are shown in Table \ref{tab:hyperparams}. All hyperparameters are shared between runs.

\begin{table}[!h]
\centering
    \begin{tabular}{l l}
    \toprule
    Hyperparameter & Value \\
    \midrule
    Max Epochs & 20 \\
    Learning Rate & 2e-6\\
    Optimizer & AdamW \\
    Learning Rate Scheduler & Cosine \\
    Warm-up Ratio & 0.1 \\
    Effective Batch Size & 256 \\
    Max Gradient Norm & 10.0 \\
    Mixed Precision & \texttt{bfloat16} \\
    \midrule
    Early Stopping Criterion & XCOMET-XL w/o Refs \\
    Early Stopping Patience & 3 \\
    Early Stopping Epsilon & 0.00001 \\
    Evaluation Frequency & Epoch \\
    Max New Tokens & 64 \\
    \midrule
    $\beta^*$ & $0.1$ \\
    Average Log-Probabilities & \texttt{True} \\
    Normalize Loss & \texttt{True} \\
    \bottomrule
    \end{tabular}
\caption{Hyperparameters for all training runs. * indicates that this parameter only affects the preference optimization techniques}
\label{tab:hyperparams}
\end{table}



\end{document}